\documentclass[letterpaper]{article} % DO NOT CHANGE THIS
\usepackage{aaai2026}  % DO NOT CHANGE THIS
\usepackage{times}  % DO NOT CHANGE THIS
\usepackage{helvet}  % DO NOT CHANGE THIS
\usepackage{courier}  % DO NOT CHANGE THIS
\usepackage[hyphens]{url}  % DO NOT CHANGE THIS
\usepackage{graphicx} % DO NOT CHANGE THIS
\usepackage{natbib}  % DO NOT CHANGE THIS AND DO NOT ADD ANY OPTIONS TO IT
\usepackage{caption} % DO NOT CHANGE THIS AND DO NOT ADD ANY OPTIONS TO IT
\usepackage{algorithm}
\usepackage{algorithmic}

\usepackage{newfloat}
\usepackage{listings}
\DeclareCaptionStyle{ruled}{labelfont=normalfont,labelsep=colon,strut=off} % DO NOT CHANGE THIS
\floatstyle{ruled}
\newfloat{listing}{tb}{lst}{}
\floatname{listing}{Listing}
\usepackage{amsfonts}
\usepackage{amsmath}
\usepackage{pifont}

\usepackage{booktabs}   % For professional-quality tables: \toprule, \midrule, \bottomrule
\usepackage{multirow}   % To span table cells across multiple rows
\usepackage{array}      % For defining new column types
\usepackage{makecell}
\usepackage[table]{xcolor}
\usepackage{nicematrix}

\usepackage{tcolorbox}
\tcbuselibrary{most}
\usepackage{alltt}

\usepackage{subcaption}
\usepackage{graphicx}   % To include images
\usepackage{tikz}       % pgfplots is built on top of tikz

\nocopyright
\title{AI-Salesman: Towards Reliable Large Language Model Driven Telemarketing}
\author {
    Qingyu Zhang\textsuperscript{\rm 1, 2}\equalcontrib,
    Chunlei Xin\textsuperscript{\rm 1, 2}\equalcontrib,
    Xuanang Chen\textsuperscript{\rm 1},
    Yaojie Lu\textsuperscript{\rm 1}\thanks{Corresponding author.},
    Hongyu Lin\textsuperscript{\rm 1}\footnotemark[2],\\
    Xianpei Han\textsuperscript{\rm 1, 2},
    Le Sun\textsuperscript{\rm 1, 2},
    Qing Ye\textsuperscript{\rm 3},
    Qianlong Xie\textsuperscript{\rm 3},
    Xingxing Wang\textsuperscript{\rm 3}
}
\affiliations {
    \textsuperscript{\rm 1}Chinese Information Processing Laboratory, Institute of Software, Chinese Academy of Sciences\\
    \textsuperscript{\rm 2}University of Chinese Academy of Sciences\\
    \textsuperscript{\rm 3}Independent Researcher
}

\usepackage{bibentry}
\begin{document}

\maketitle

\begin{abstract}
Goal-driven persuasive dialogue, exemplified by applications like telemarketing, requires sophisticated multi-turn planning and strict factual faithfulness, which remains a significant challenge for even state-of-the-art Large Language Models (LLMs). A lack of task-specific data often limits previous works, and direct LLM application suffers from strategic brittleness and factual hallucination. In this paper, we first construct and release \textbf{TeleSalesCorpus}, the first real-world-grounded dialogue dataset for this domain. We then propose AI-Salesman, a novel framework featuring a dual-stage architecture. For the training stage, we design a Bayesian-supervised reinforcement learning algorithm that learns robust sales strategies from noisy dialogues. For the inference stage, we introduce the Dynamic Outline-Guided Agent (DOGA), which leverages a pre-built script library to provide dynamic, turn-by-turn strategic guidance. Moreover, we design a comprehensive evaluation framework that combines fine-grained metrics for key sales skills with the LLM-as-a-Judge paradigm. Experimental results demonstrate that our proposed AI-Salesman significantly outperforms baseline models in both automatic metrics and comprehensive human evaluations, showcasing its effectiveness in complex persuasive scenarios.\footnote{The \textbf{TeleSalesCorpus} is available at \url{https://huggingface.co/datasets/ICIP/TeleSalesCorpus}.} 
\end{abstract}

%     \link{Datasets}{https://huggingface.co/datasets/ICIP/TeleSalesCorpus}

\section{Introduction}

While conversational AI has made significant strides in both structured task-oriented dialogue \citep{ham-etal-2020-end, hosseini2020simple, xu-etal-2024-rethinking} and unconstrained open-domain chit-chat \citep{gao-etal-2018-neural-approaches, roller-etal-2021-recipes, friedman-etal-2025-representing}, a critical and challenging frontier remains underexplored: goal-driven persuasive dialogue for intelligent marketing, unlike conventional dialogue tasks, intelligent marketing, exemplified by telemarketing, requires conversational AI to actively strategize, persuade, and guide users toward specific outcomes. This presents a unique confluence of high-stakes challenges that current large language models (LLMs) struggle to address effectively.

The core challenges of intelligent telemarketing are threefold.
First is the challenge of satisfaction. The AI must not only generate human-like responses but also navigate a wide variety of marketing scenarios, each with its own complex strategies and logical flows. General-purpose large language models, despite their fluency, struggle to capture and reliably execute these diverse, long-horizon conversational plans \citep{valmeekam2023on, pan2025why, chen2025consistentchat,lin2025speculativedecodingreimaginedmultimodal}, failing to satisfy the strategic requirements of the task. Second is the challenge of faithfulness. In high-stakes sales interactions, the AI must adhere strictly to the constraints of the product or service. However, the propensity of LLMs for factual hallucination \citep{maynez-etal-2020-faithfulness, rawte-etal-2023-troubling, atanasova-etal-2023-faithfulness, chen2025xiangqir1enhancingspatialstrategic,chen-etal-2025-think} poses an unacceptable risk, potentially resulting in misleading claims or inaccurate commitments. Third is the challenge of customization. Each customer possesses a unique background, with distinct concerns and points of interest. Effective persuasion requires tailoring arguments and information delivery to individual needs. Yet, LLMs frequently produce generic responses and lack the strategic reasoning necessary to address specific objections effectively \citep{fu2023improvinglanguagemodelnegotiation}.

\begin{figure*}[t]
    \centering
    \includegraphics[width=0.95\textwidth]{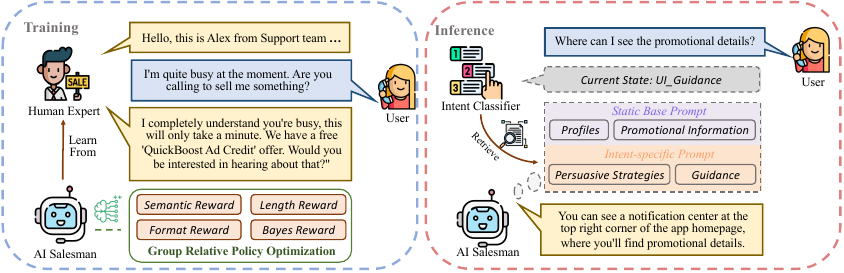} 
    \caption{Overview of Training and Inference for the AI Salesman.}
    \label{Overview_framework}
\end{figure*}

To address these multifaceted challenges, this paper introduces AI-Salesman, an end-to-end framework that tackles these issues through innovations at both the training and inference stages, as illustrated in Figure~\ref{Overview_framework}. Specifically, AI-Salesman integrates two core mechanisms to achieve this.
First, to satisfy the critical demands of satisfaction and faithfulness, we introduce a novel reward function grounded in Bayesian principles into our Group Relative Policy Optimization (GRPO) training process \citep{shao2024deepseekmath}. Moving beyond conventional outcome-based rewards, our approach directly supervises the model’s intermediate reasoning. Inspired by Bayesian principles, we decompose the reward signal for a thought process into two intuitive criteria: a prior that captures the intrinsic coherence of the reasoning itself, and a likelihood that measures its strategic utility in justifying the expert's final response. By optimizing for both coherent reasoning and effective outcomes, the model learns to generate responses that are both factually grounded and persuasive, thereby enhancing user satisfaction and faithfulness.
Second, to enable customization, we propose the Dynamic Outline-Guided Agent (DOGA), a framework that operates during the inference stage. To overcome the generic responses common with static prompting, DOGA dynamically constructs a tailored strategy outline for each turn. By analyzing the user’s profile, real-time intent, and dialogue history, it retrieves the most relevant persuasive strategies from a pre-verified library. This curated outline then guides the LLM, ensuring its responses are strategically targeted to each customer's unique concerns and objections.

Unfortunately, a significant barrier to progress in this domain is the absence of specialized training data and effective evaluation methods for telemarketing \citep{he-etal-2018-decoupling,wang-etal-2019-persuasion}. To address this gap, we first introduce TeleSalesCorpus, a large-scale corpus of high-fidelity dialogues generated through a state-aware simulation grounded in real-world expert interactions. This corpus captures the complex patterns, customer objections, and conversational nuances characteristic of authentic sales conversations. Second, moving beyond simplistic success metrics, we propose a comprehensive evaluation framework specifically designed for telemarketing to enable fine-grained analysis. To systematically assess a model's ability to achieve strategic satisfaction, maintain factual faithfulness, and deliver persuasive customization, we define six sales capabilities, ranging from Business Analysis to Objection Handling, each assessed using a detailed rubric composed of seven qualitative metrics. By integrating this structured evaluation schema with the LLM-as-a-Judge paradigm \citep{zheng2023judging, chan2024chateval}, our framework supports rigorous and comprehensive assessment of model performance across diverse scenarios. This evaluation approach provides a scalable offline alternative to resource-intensive online A/B tests.

Overall, our contributions can be summarized as follows:
\begin{itemize}
    \item We propose AI-Salesman, a novel end-to-end framework that integrates reasoning-aware reinforcement learning with dynamic outline-guided inference. To the best of our knowledge, this is the first LLM-based framework specifically designed for real-world telemarketing that systematically addresses the challenges of satisfaction, faithfulness, and customization.
    \item We construct and release TeleSalesCorpus, the first large-scale, high-fidelity dialogue dataset grounded in real-world sales conversations, specifically designed for training and evaluating telemarketing models.
    \item We propose a comprehensive offline evaluation framework across six core sales capabilities, enabling efficient and rigorous assessment of models’ practical sales proficiency in diverse scenarios.
\end{itemize}

\section{Telemarketing Scenarios}

To systematically analyze model performance in telemarketing, this section addresses two key aspects. First, we formally define the dialogue generation task to articulate its underlying structure. Second, we introduce a comprehensive framework designed to evaluate the model performance across critical sales capabilities and qualitative metrics.

\subsection{Task Definition}

We model telemarketing dialogue as a conditionally constrained sequence generation task. At each turn $t$, the model generates a response based on the system prompt $\mathcal{P}$ and the dialogue history $\mathcal{H}_t = \mathcal{H}_{t-1} \oplus U_t$, where $U_t$ is the user's utterance at turn t. The prompt $\mathcal{P}$ defines the task's global context, including a set of goals $G = \{g_1, \dots, g_n\}$ and constraints $C = \{c_1, \dots, c_m\}$.

The model's objective is to generate a response sequence $A_t$ that maximizes its conditional probability given the inputs $(\mathcal{P}, \mathcal{H}_t)$. Formally, we seek the optimal response $A_t^*$:
\begin{equation}
    A_t^* = \arg\max_{A_t \in \mathcal{V}^*} P(A_t | \mathcal{P}, \mathcal{H}_t)
\end{equation}
where $\mathcal{V}$ is the model's vocabulary and $\mathcal{V}^*$ denotes its Kleene closure, representing the set of all possible sequences the model can generate.

This generation is subject to two primary conditions. First, the response $A_t$ must adhere to all predefined rules, such that for every constraint $c \in C$, the condition $c(A_t) = 1$ is satisfied. Second, the response must be goal-oriented, designed to maximize the expectation of achieving the final task goals defined in $G$.

\subsection{Evaluation Framework}
\label{sec:evaluation_framework}
Our evaluation framework is built upon two core components: six fundamental sales capabilities required for the task and a rubric of seven evaluation metrics for granular, turn-by-turn assessment. Detailed descriptions of these components are provided in Appendix~\ref{app:framework_details}.

The six capabilities cover the entire lifecycle of a sales call: Role-playing, Business Analysis, Activity Introduction, Idle-chat Rejection, Objection Handling, and Operational Guidance. To provide a fine-grained assessment across these capabilities, we evaluate each response using seven qualitative metrics: Guideline Adherence(\textbf{Gui.}), Factual Correctness(\textbf{Fac.}), Logical Coherence(\textbf{Log.}), User Need Fulfillment(\textbf{Use.}), Response Richness(\textbf{Res.}), Safety(\textbf{Saf.}), and Completeness(\textbf{Com.}).

To operationalize this framework at scale, we employ GPT-4 as a judge. For each dialogue, the LLM-judge is given the conversation history, ground-truth data, and our metric definitions. Then it synthesizes these inputs to generate a holistic quality score on a 1-10 scale. This approach enables nuanced, context-aware evaluation that approximates human judgment for robustly benchmarking different models.

\section{End-to-End Intelligent Sales System}

\begin{figure}[t!]
    \centering
    \includegraphics[width=0.48\textwidth]{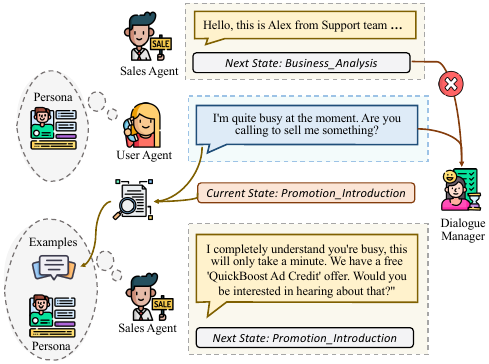}
    \caption{Data Construction Framework Overview.}
    \label{fig:data_construction_framework}
\end{figure}

\subsection{Data Construction}
\label{sub:data_construction}
The availability of suitable training data fundamentally constrains the development of a robust, goal-oriented persuasive dialogue system. Existing datasets \citep{wang-etal-2019-persuasion,he-etal-2018-decoupling} do not adequately address the unique challenges of telemarketing, such as complex business rules and specific promotional objectives. To bridge this gap, we constructed TeleSalesCorpus, a dataset using a semi-synthetic framework that leverages real-world expertise to generate high-fidelity, goal-oriented dialogues.

Our data creation process employs a state-aware, three-agent simulation, as illustrated in the Figure~\ref{fig:data_construction_framework} provided. The framework features a User Agent with a distinct persona, a Sales Agent responsible for persuasion, and a central Dialogue Manager that orchestrates the interaction. At each turn, when the User Agent responds, the Dialogue Manager intervenes. It first adjudicates the true conversational state, overriding incorrect state predictions from the sales agent. Then, it queries a pre-compiled library of real-world interaction examples, retrieving a strategically relevant example based on the current state. This example is used to dynamically guide the Sales Agent in crafting a response that is both contextually appropriate and strategically sound.

This process is grounded in assets distilled from real dialogues and diverse, LLM-authored business scenarios. Following a rigorous, multi-faceted quality assurance protocol, our pipeline produced a final dataset of 2,000 high-fidelity conversations. The detailed methodology for each stage—asset distillation, dialogue simulation, and quality assurance—is provided in Appendix~\ref{app:data_construction_details}.

\begin{figure*}[ht]
    \centering
    \includegraphics[width=0.99\textwidth]{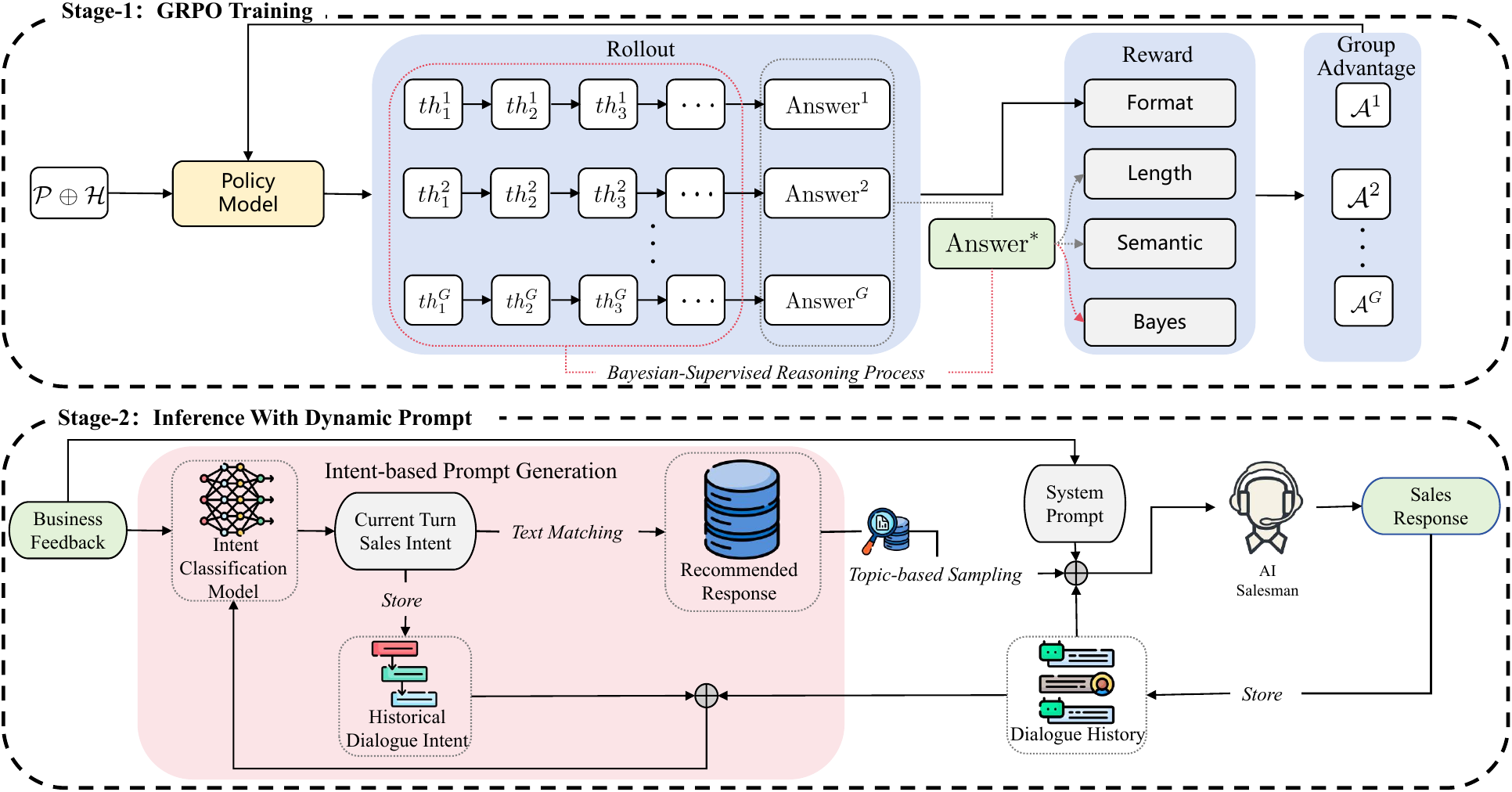}
    \caption{AI Salesman Framework Overview.}
    \label{fig:ai_saleman_framework}
\end{figure*}

\subsection{Stage-1: GRPO Training}
To address the core challenges of satisfaction and faithfulness in intelligent telemarketing, we propose a policy optimization framework that synergizes the Group Relative Policy Optimization (GRPO) algorithm\citep{shao2024deepseekmath} with a novel Bayesian-Supervised Reasoning reward. GRPO facilitates online exploration of sales strategies, enabling the model to learn robust policies from noisy data. This exploration is guided by our Bayesian reward, which uniquely assesses the model's intermediate reasoning process. It assigns a higher value to reasoning that provides a logically sound and factually grounded justification for the final response. This core signal is supplemented by several auxiliary rewards designed to maintain structural and semantic integrity. By optimizing this reward via GRPO, the model learns to generate responses that are both persuasive, to enhance Satisfaction, and factually accurate, to ensure Faithfulness.

As illustrated in Figure~\ref{fig:ai_saleman_framework}, our end-to-end training is driven by the GRPO algorithm. For the $t$-th turn given input $\mathcal{P} \oplus \mathcal{H}$, the model first performs $G$ parallel rollouts to generate a group of candidate sequences $\{A_t^{(i)}\}_{i=1}^G$. The algorithm then uses the reward signal $R^{(i)}$ from each sequence to compute a normalized group advantage score, $\mathcal{A}^{(i)}$, and subsequently updates the policy model. The details of the GRPO algorithm are provided in Appendix~\ref{app:appendix_grpo}.

\subsection*{Reward Function Design}
The total reward $R$ is a weighted sum of four components, evaluating different aspects of the generated sequence $A_t^{(i)}$ against the ground-truth reference $A_t^*$:
\begin{equation}
R(A_t^{(i)}, A_t^*) = \sum_{k \in \{\text{bayes, format, len, sem}\}} w_k R_k(A_t^{(i)}, A_t^*)
\end{equation}
where $w_k$ are hyperparameter weights.

\subsection*{Core Reward}
\paragraph{Bayesian-Supervised Reasoning ($R_{\text{bayes}}$)}
This reward guides the model's internal reasoning chain, $Th_t$. Grounded in Bayesian principles, our objective is to align this reasoning chain with the reference answer $A_t^*$ by maximizing their joint probability, $P(Th_t, A_t^*)$. Accordingly, the reward is defined as the log-joint probability, which decomposes into two terms estimated by the model $\pi_\theta$ itself. The detailed theoretical derivation is provided in Appendix~\ref{app:bayesian_theorem}.
\begin{equation}
\label{eq:bayes_reward_main}
\begin{split}
R_{\text{bayes}}(Th_t^{(i)}, A_t^*) &= 
\underbrace{\sum_{j=1}^{m} \log \pi_\theta(th_j^{(i)} \mid \hat{\mathcal{P}}, th_{<j}^{(i)})}_{\text{Prior: Reasoning Fluency}} \\
&+
\underbrace{\sum_{k=1}^{n} \log \pi_\theta(y_k^* \mid \hat{\mathcal{P}}, Th_t^{(i)}, y_{<k}^*)}_{\text{Likelihood: Reasoning Utility}}
\end{split}
\end{equation}
where $\hat{\mathcal{P}}$ is the shared context.

\subsection*{Auxiliary Reward}
\paragraph{Format Adherence ($R_{\text{format}}$)}
A reward that ensures the output follows the predefined \verb|"<think>...</think><answer>...</answer>"| schema.
\begin{equation}
    R_{\text{format}}(A_t^{(i)}) = f_{\text{format}}(A_t^{(i)})
\end{equation}
where $f_{\text{format}}(\cdot)$ is a function that yields 1 if the sequence $A_t^{(i)}$ conforms to the required schema, and 0 otherwise.

\paragraph{Relative Length Consistency ($R_{\text{len}}$)}
This aims to align the output length with the reference answer by penalizing the squared relative deviation from the target length $L(A_t^*)$.
\begin{equation}
    R_{\text{len}}(A_t^{(i)}, A_t^*) =1 - \left( \frac{|L(A_t^{(i)}) - L(A_t^*)|}{L(A_t^*)}\right)^2
\end{equation}

\paragraph{Semantic Similarity ($R_{\text{sem}}$)}
To measure semantic alignment, we compute the cosine similarity $s^{(i)}$ between the generated and reference answers using a sentence-embedding model. The score is normalized against a baseline similarity $s_{\text{base}}$ for a more robust signal.
\begin{equation}
    R_{\text{sem}}(A_t^{(i)}, A_t^*) = \frac{s^{(i)} - s_{\text{base}}}{1 - s_{\text{base}} + \epsilon}
\end{equation}

\begin{table*}[t]
    \centering
    \begin{NiceTabular*}{\textwidth}{l l @{\extracolsep{\fill}} c c c c c c c c}
        \toprule
        \textbf{Capability} & \textbf{Model} & \textbf{Mean} & \textbf{Gui.} & \textbf{Fac.} & \textbf{Log.}  & \textbf{Use.} & \textbf{Res.} & \textbf{Saf.} & \textbf{Com.} \\
        \midrule
        \Block{4-1}{\textbf{Role-playing}}  
        & Baseline & 5.54 & \underline{4.83} & 5.70 & 5.94 & 5.40 & 5.02 & 7.20 & 4.68  \\
        & SFT-only    & 5.66 & 4.78 & 5.90 & 6.05 & 5.61 & \underline{5.08} & 7.27 & 4.92 \\
        & GRPO w/ SFT & \underline{5.75} & 4.79 & \underline{5.95} & \underline{6.16} & \underline{5.72} & \underline{5.08} & \underline{7.41} & \underline{5.13} \\
        & \cellcolor{blue!8}\textbf{Ours} & \cellcolor{blue!8}\textbf{6.31} & \cellcolor{blue!8}\textbf{5.81} & \cellcolor{blue!8}\textbf{6.5} & \cellcolor{blue!8}\textbf{6.62} & \cellcolor{blue!8}\textbf{6.16} & \cellcolor{blue!8}\textbf{5.76} & \cellcolor{blue!8}\textbf{7.6} & \cellcolor{blue!8}\textbf{5.75} \\
        
        \midrule
        \Block{4-1}{\textbf{Business Analysis}}  
        & Baseline & 6.49 & 5.42 & 6.44 & 7.05 & 6.67 & 6.19 & 7.72 & 5.91 \\
        & SFT-only    & 6.78 & 5.39 & 7.15 & \underline{7.24} & \underline{7.04} & 6.39 & 7.83 & 6.41 \\
        & GRPO w/ SFT & \underline{6.86} & \underline{5.51} & \underline{7.39} & \underline{7.24} & 6.97 & \underline{6.59} & \underline{7.88} & \underline{6.44} \\
        & \cellcolor{blue!8}\textbf{Ours} & \cellcolor{blue!8}\textbf{7.40} & \cellcolor{blue!8}\textbf{5.96} & \cellcolor{blue!8}\textbf{7.87} & \cellcolor{blue!8}\textbf{7.78} & \cellcolor{blue!8}\textbf{7.61} & \cellcolor{blue!8}\textbf{7.23} & \cellcolor{blue!8}\textbf{7.94} & \cellcolor{blue!8}\textbf{7.43} \\
        
        \midrule
        \Block{4-1}{\textbf{Activity Introduction}}  
        & Baseline & 5.91 & \underline{5.39} & 5.28 & 6.43 & 6.18 & \underline{5.76} & 7.39 & 4.97 \\
        & SFT-only & 5.86 & 5.16 & 5.32 & 6.38 & \underline{6.23} & 5.56 & 7.33 & 5.04 \\
        & GRPO w/ SFT & \underline{5.94} & 5.08 & \underline{5.41} & \underline{6.49} & 6.15 & 5.62 & \underline{7.45} & \underline{5.36} \\
        & \cellcolor{blue!8}\textbf{Ours} & \cellcolor{blue!8}\textbf{6.75} & \cellcolor{blue!8}\textbf{6.55} & \cellcolor{blue!8}\textbf{5.98} & \cellcolor{blue!8}\textbf{7.13} & \cellcolor{blue!8}\textbf{7.07} & \cellcolor{blue!8}\textbf{6.71} & \cellcolor{blue!8}\textbf{7.94} & \cellcolor{blue!8}\textbf{5.86} \\
        
        \midrule
        \Block{4-1}{\textbf{Idle-chat Rejection}}  
        & Baseline & 4.66 & \underline{4.36} & 4.35 & 5.10 & 4.48 & 4.41 & 6.31 & 3.59 \\
        & SFT-only & 4.86 & 3.96 & 4.68 & 5.36 & 4.83 & \underline{4.67} & 6.63 & 3.89 \\
        & GRPO w/ SFT & \underline{4.95} & 4.11 & \underline{4.72} & \underline{5.48} & \underline{4.90} & 4.59 & \underline{6.78} & \underline{4.09} \\
        & \cellcolor{blue!8}\textbf{Ours} & \cellcolor{blue!8}\textbf{5.73} & \cellcolor{blue!8}\textbf{5.49} & \cellcolor{blue!8}\textbf{5.52} & \cellcolor{blue!8}\textbf{6.19} & \cellcolor{blue!8}\textbf{5.59} & \cellcolor{blue!8}\textbf{5.50} & \cellcolor{blue!8}\textbf{6.99} & \cellcolor{blue!8}\textbf{4.81} \\
        
        \midrule
        \Block{4-1}{\textbf{Objection Handling}}  
        & Baseline & 4.77 & 4.60 & 3.92 & 5.18 & 4.97 & 4.47 & 6.46 & 3.80 \\
        & SFT-only & 5.24 & 5.19 & \underline{4.64} & 5.75 & \underline{5.22} & 5.01 & 6.56 & 4.34 \\
        & GRPO w/ SFT & \underline{5.33} & \underline{5.41} & 4.58 & \underline{5.82} & 5.09 & \underline{5.23} & \underline{6.69} & \underline{4.49} \\
        & \cellcolor{blue!8}\textbf{Ours} & \cellcolor{blue!8}\textbf{6.00} & \cellcolor{blue!8}\textbf{6.24} & \cellcolor{blue!8}\textbf{4.65} & \cellcolor{blue!8}\textbf{6.57} & \cellcolor{blue!8}\textbf{6.22} & \cellcolor{blue!8}\textbf{6.02} & \cellcolor{blue!8}\textbf{7.49} & \cellcolor{blue!8}\textbf{4.82} \\
        
        \midrule
        \Block{4-1}{\textbf{Operational Guidance}}  
        & Baseline & 5.39 & 4.44 & 6.13 & 5.52 & 5.33 & 4.68 & 6.56 & 5.09 \\
        & SFT-only & 5.71 & 4.84 & 6.15 & \underline{5.87} & 5.54 & \underline{5.29} & 6.99 & 5.29 \\
        & GRPO w/ SFT & \underline{5.78} & \underline{4.90} & \underline{6.20} & 5.81 & \underline{5.68} & 5.16 & \underline{7.18} & \underline{5.51} \\
        & \cellcolor{blue!8}\textbf{Ours} & \cellcolor{blue!8}\textbf{6.74} & \cellcolor{blue!8}\textbf{6.26} & \cellcolor{blue!8}\textbf{7.33} & \cellcolor{blue!8}\textbf{6.71} & \cellcolor{blue!8}\textbf{6.50} & \cellcolor{blue!8}\textbf{6.09} & \cellcolor{blue!8}\textbf{7.63} & \cellcolor{blue!8}\textbf{6.67} \\
        \bottomrule
    \end{NiceTabular*}%
    \caption{Performance comparison of different training pipelines. Our framework significantly outperforms all competing baselines. The top-performing model, Ours, utilizes direct reinforcement learning, bypassing the SFT stage. Best results in each block are in \textbf{bold}. The second-best results in each block are \underline{underlined}.}
    \label{tab:main_results}
\end{table*}

\subsection{Stage-2: Inference With Dynamic Prompt}
We propose the Dynamic Outline-Guided Agent (DOGA) to enable customization in telemarketing by overcoming the rigidity of static prompts. Our framework decouples high-level strategy from turn-level execution by generating turn-specific guidance from a pre-structured script library. This process is composed of two stages: an offline library construction phase and a real-time dynamic prompt assembly pipeline. This structure ensures that model responses are personalized and contextually appropriate. More details of the DOGA framework are detailed in Appendix~\ref{app:doga_details}.

\subsubsection{Offline Stage: Structured Script Library Construction}
The foundation of our framework is a high-quality library of sales scripts and templates. This library is created offline by extracting, clustering, and summarizing effective strategies from a corpus of successful historical dialogues. This process distills best practices into a reusable resource indexed by dialogue intent.

\subsubsection{Online Stage: Real-time Dialogue Management}
During a live conversation, DOGA employs the real-time pipeline shown in Figure~\ref{fig:ai_saleman_framework}. At each turn, an Intent Classification Model first predicts the user's current turn sales intent. This intent is used to retrieve a relevant recommended response template from our pre-built library. Finally, this turn-specific guidance is combined with the system prompt and the full dialogue history to assemble a dynamic system prompt. This prompt steers the model to generate a response that is strategically aligned with the immediate conversational goal.

\section{Experiments}
This section presents a series of experiments designed to evaluate the effectiveness of our proposed AI-Salesman framework. We first detail the experimental setup, including the datasets, models, and evaluation protocols. We then present the main results comparing our full method against several baselines. Finally, through extensive ablation studies, scalability analysis, and human evaluations, we validate the contributions of the key components of our framework.

\subsubsection{Datasets}
We utilize two datasets with distinct roles in our experiments:
\begin{itemize}
    \item \textbf{TeleSalesCorpus (Syn-Data)}: To ensure the reproducibility and openness of our research, we introduce this synthetic dataset, which will be made publicly available. Constructed as described in Section~\ref{sub:data_construction}, it contains 2,000 high-fidelity, multi-turn dialogues. 
    
    \item \textbf{Real-world Tele-sales Dataset (Real-Data)}: This is our dataset for large-scale training. It consists of over 8,000 real-world tele-sales dialogues. This proprietary dataset reflects the complexities of authentic sales conversations, including significant conversational noise and diversity. It is instrumental for assessing our model's performance and scalability in a realistic application setting.
\end{itemize}
\subsection{Experimental Setup}

\begin{table}[t!]
\centering
    \begin{tabular}{l c c l}
    \toprule
    \textbf{Model} & \textbf{SFT} & \textbf{GRPO} & \textbf{Inference Strategy} \\
    \midrule
    Baseline        & \ding{55}     & \ding{55}     & Few-shot \\
    SFT-only        & \ding{51}     & \ding{55}     & Few-shot \\
    GRPO w/ SFT     & \ding{51}     & \ding{51}     & DOGA \\
    \textbf{Ours}   & \ding{55}     & \ding{51}     & DOGA \\
    \bottomrule
    \end{tabular}
\caption{
    Configurations for the different models in our experiments. 
    The base model for all versions is \textbf{Qwen2.5-7B-Instruct}.
    \ding{51} indicates the stage was applied, while \ding{55} indicates it was skipped. 
}
\label{tab:model_configurations}
\end{table}

\begin{figure*}[t!]
    \centering
    %--------------------------------------------------------------------
    % Subfigure (a): Bayesian Reward Effect
    %--------------------------------------------------------------------
    \begin{subfigure}[b]{0.32\textwidth}
        \centering
        \includegraphics[width=\textwidth]{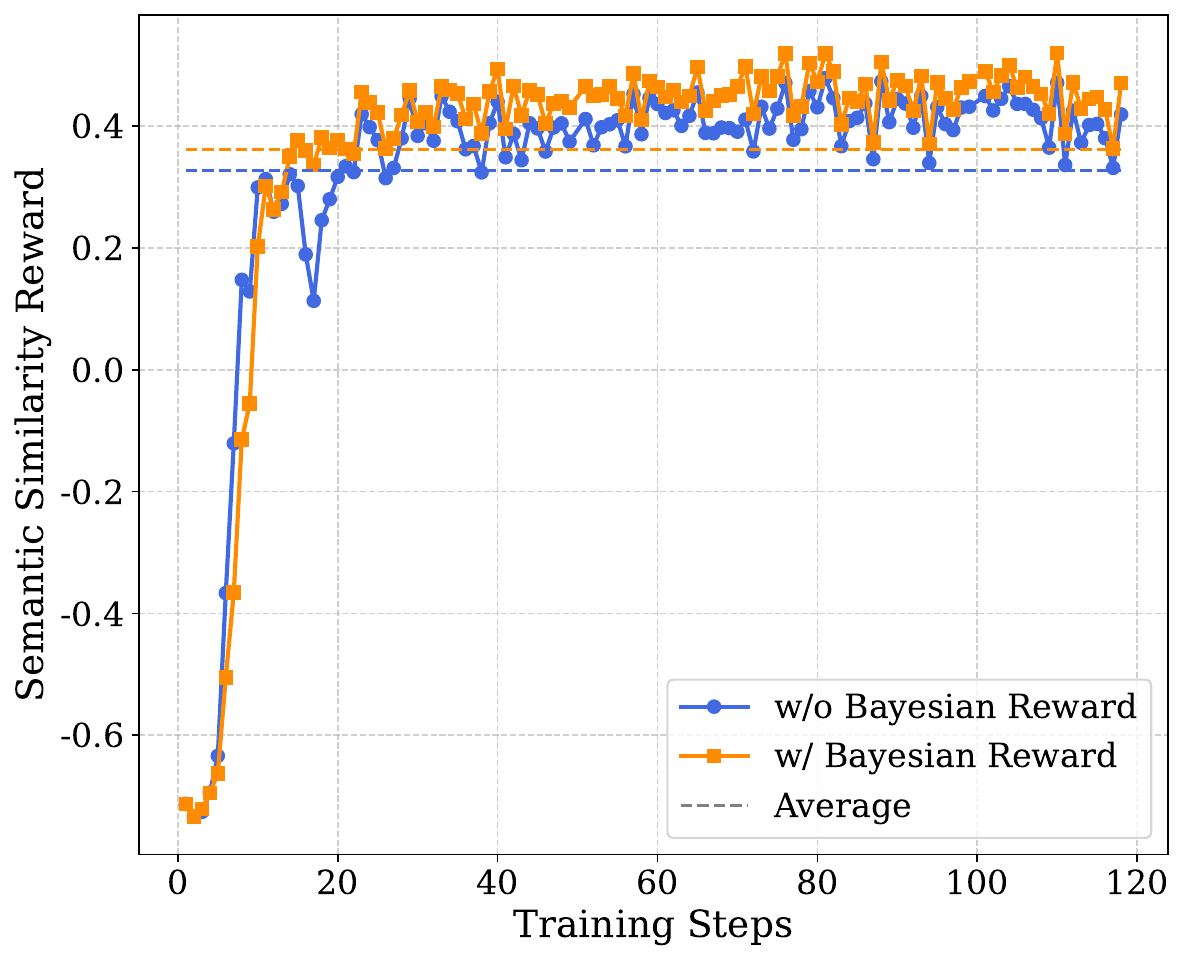} 
        \caption{Effect of $R_{\text{bayes}}$ on Syn-Data.}
        \label{fig:bayes_reward_effect_syn}
    \end{subfigure}
    \hfill % Adds horizontal space between figures
    %--------------------------------------------------------------------
    % Subfigure (b): DOGA Impact
    %--------------------------------------------------------------------
    \begin{subfigure}[b]{0.32\textwidth}
        \centering
        \includegraphics[width=\textwidth]{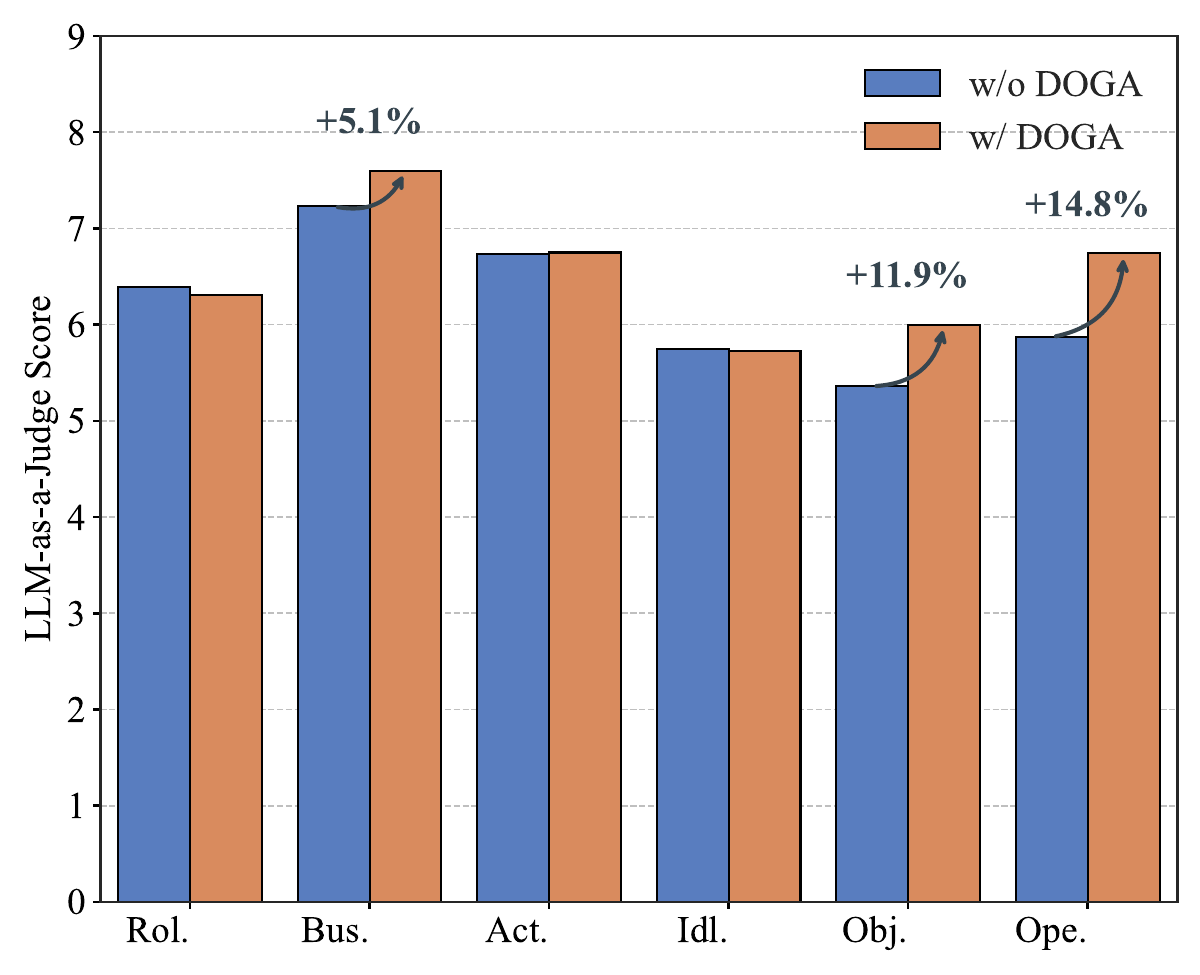} 
        \caption{Advantage in strategic capabilities.}
        \label{fig:doga_effect_nuanced}
    \end{subfigure}
    \hfill % Adds horizontal space
    %--------------------------------------------------------------------
    % Subfigure (c): Scalability
    %--------------------------------------------------------------------
    \begin{subfigure}[b]{0.32\textwidth}
        \centering
        \includegraphics[width=\textwidth]{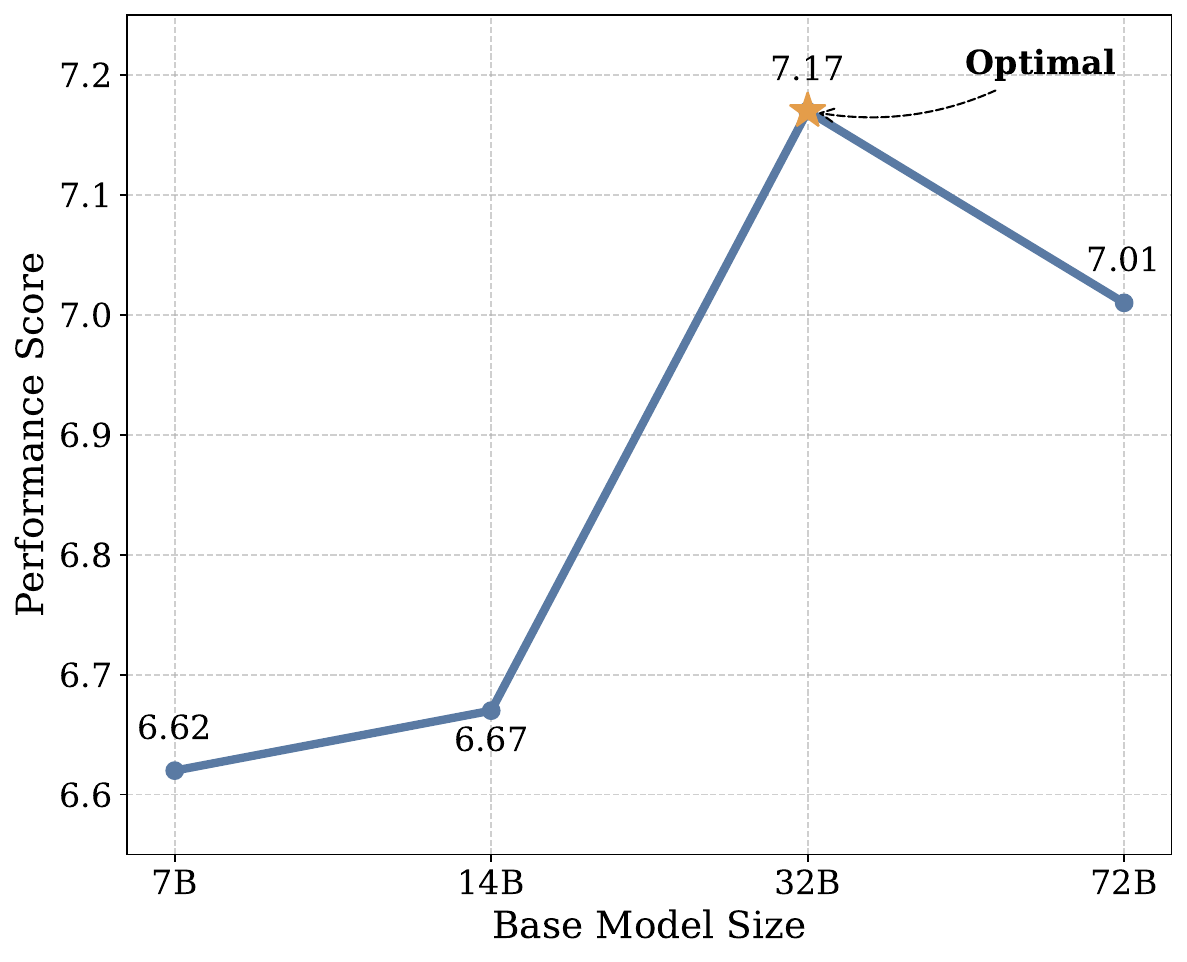} 
        \caption{Performance scaling with model size.}
        \label{fig:our_method_scale}
    \end{subfigure}
    \caption{Key experimental results. (a) Bayesian reward ($R_{\text{bayes}}$) stably raises the upper bound of the semantic similarity reward. (b) DOGA shows decisive advantages in complex, strategic capabilities. (c) Our method's performance scales effectively, with the 32B model offering an optimal trade-off.}
    \label{fig:main_experimental_results}
\end{figure*}

\subsubsection{Models and Baselines}
Our main experiment is based on the Qwen2.5-7B-Instruct model \citep{qwen25technicalreport}. As detailed training and inference configuration in Table~\ref{tab:model_configurations}. We establish a performance reference using the original \textbf{Baseline} and a standard Supervised Fine-Tuning \textbf{SFT-only model}. Our primary contribution is \textbf{Ours}, which applies the GRPO algorithm with the reward function we designed directly to the baseline. To investigate whether SFT is a necessary step for effective preference alignment, we also train a \textbf{GRPO w/ SFT} model by applying GRPO after the SFT stage.

\subsubsection{Evaluation Metrics}
As detailed in Section~\ref{sec:evaluation_framework}, we use the LLM-as-a-Judge paradigm with GPT-4 as the evaluator. Each dialogue turn is scored from 1 to 10 across seven metrics. The final score for each of the six core sales capabilities is the arithmetic mean of seven metrics.

\subsection{Main Results}
The comprehensive performance evaluation, detailed in Table~\ref{tab:main_results}, empirically substantiates the remarkable efficacy of our proposed training paradigm. Our final model, denoted \textbf{Ours}, establishes a new state-of-the-art, achieving dominant scores across the vast majority of capabilities and dimensions evaluated. Our analysis reveals three principal findings:

\begin{itemize}
    \item \textbf{Finding 1: Domain-specific SFT establishes a robust but limited performance baseline.} 
    The results indicate a mixed but overall positive effect from SFT. This confirms its role as a preliminary adaptation stage. While SFT led to significant gains in areas like Business Analysis (6.49 → 6.78) and Objection Handling (4.77 → 5.24), its impact on more complex skills was limited. For example, the score for Role-playing grew minimally from 5.54 to 5.66. This demonstrates that SFT is effective at mimicking explicit patterns but struggles with tasks requiring deeper strategic generalization.
    
    \item \textbf{Finding 2: SFT creates a performance bottleneck for reinforcement learning.} 
    Our experiments show that applying reinforcement learning to an SFT-initialized model (GRPO w/ SFT) offers negligible performance gain over the SFT model alone, with the overall mean score across all capabilities only increasing minimally from 5.69 (SFT-only) to 5.77 (GRPO w/ SFT). We conclude that SFT, by forcing the model to mimic a noisy and suboptimal dataset, traps its policy in a narrow, flawed space. 
    This severely restricts RL's ability to explore and discover superior strategies, resulting in a final policy that fails to meaningfully diverge from the flawed behaviors learned during SFT. The model thus adheres to rules but lacks conversational richness.

    \item \textbf{Finding 3: Direct RL optimization without SFT unlocks superior performance.} 
    In stark contrast, optimizing a base model directly with our GRPO reward signal yields a holistically superior model, boosting the overall mean score from the Baseline's 5.46 to 6.49—a significant 18.9\% increase. By being liberated from the constraints of imitating a potentially suboptimal reference corpus, the model learns to internalize the underlying business logic and knowledge directly from rewards. This approach achieves high performance across all dimensions—excelling not only in Richness (Res.) and User Satisfaction (Use.) but also maintaining strong Guideline Adherence (Gui.), proving it's a more effective path to developing a capable and adaptive sales model.
\end{itemize}

\subsection{Ablation Studies}
To evaluate the specific contributions of our proposed components, we conducted a series of ablation studies. These experiments are designed to isolate and quantify the impact of our reward functions and DOGA.

\paragraph{Quantitative Analysis of Reward Components}
We first investigated the individual importance of the key signals in our composite reward function. To do this, we trained two ablated versions of our model:
\begin{itemize}
    \item \textbf{GRPO w/o $R_{\text{bayes}}$}: The model was trained without the Bayesian-Supervised Reasoning Reward, removing the explicit supervision on the internal thought process.
    \item \textbf{GRPO w/o $R_{\text{sem}}$}: The model was trained without the Semantic Similarity Reward, removing the direct pressure to align the final answer with the expert reference.
\end{itemize}

\begin{table}[t!]
\centering
    \begin{tabular}{l c}
    \toprule
    \textbf{Model Version}  & \textbf{Mean Score} \\
    \midrule
    GRPO w/o $R_{\text{bayes}}$ & 6.15 \\
    GRPO w/o $R_{\text{sem}}$   & 6.39 \\
    \multicolumn{1}{c}{\textbf{Ours}}     & \textbf{6.49} \\ 
    \bottomrule
    \end{tabular}
\caption{Ablation study of reward components. The LLM-as-a-Judge calculates the mean score across all evaluation capabilities.}
\label{tab:ablation_reward}
\end{table}

As shown in Table~\ref{tab:ablation_reward}, the results clearly demonstrate the criticality of both components. Removing the Bayesian reward ($R_{\text{bayes}}$) led to a 5.2\% drop in the mean score, while removing the semantic reward ($R_{\text{sem}}$) caused a 1.5\% decrease. This confirms that both reward signals are essential for guiding the model. $R_{\text{sem}}$ directly optimizes for output quality, while $R_{\text{bayes}}$ ensures the underlying reasoning is sound, which indirectly but powerfully contributes to the generation of high-quality and reliable responses.

\paragraph{Visualizing the Effect of Bayesian Reward}
To visualize the effect of our most novel component, the Bayesian reward, we plotted the training-time semantic similarity reward on our synthesized dataset, TeleSalesCorpus (Syn-Data). As shown in Figure~\ref{fig:bayes_reward_effect_syn}, the model trained with $R_{\text{bayes}}$ converges to a higher semantic similarity ceiling steadily. This suggests that by penalizing illogical thought processes, the Bayesian reward acts as an internal verifier, preventing the model from exploring ineffective generation paths and steering it more directly toward producing answers that are semantically aligned with expert behavior. The reward demonstrates a similar effect on a challenging real-world dataset, as detailed in the appendix~\ref{app:real_data_results}.

\paragraph{Effectiveness of DOGA}

A comparative analysis of our DOGA framework against a static prompt on six sales capabilities reveals two key findings (Figure~\ref{fig:doga_effect_nuanced}):

\begin{itemize}
    \item \textbf{Finding 1: DOGA excels in complex tasks.} 
    It achieved significant performance gains in Business Analysis (+4.9\%), Objection Handling (+11.1\%), and Operational Guidance (+14.7\%). This performance boost is driven by its ability to dynamically adapt, drawing from a library of expert templates to deliver more detailed and accurate contextual guidance in real-time, surpassing the limitations of static prompts.
    \item \textbf{Finding 2: A trade-off exists between strategic precision and conversational naturalness.}The static prompt performed marginally better in Role-playing and Idle-chat Rejection. DOGA's template injection, while precise, can sound formulaic. For simple tasks, the static prompt's direct rules are more efficient than DOGA's complex retrieval cycle.
\end{itemize}

In conclusion, DOGA is a specialized instrument, not a universal upgrade. Its primary value is enhancing strategic reasoning and procedural adherence in complex, goal-oriented dialogues, making it indispensable for developing sophisticated AI-Salesman.

\subsection{Scalability Analysis}
To systematically evaluate the scalability of our proposed method, we conducted a scaling experiment using the Qwen2.5-Instruct series of models, which includes variants with 7B, 14B, 32B, and 72B parameters. Each model was trained and subsequently evaluated on our curated Real-Data set. The results are shown in Figure~\ref{fig:our_method_scale}. We observed a non-linear performance trend with several key findings (detailed experimental settings are provided in Appendix~\ref{app:exp_settings}):
\begin{itemize}
\item \textbf{Marginal Gain:} Scaling from 7B to 14B yields only a minor improvement.
\item \textbf{Peak Performance:} The 32B model achieves a significantly higher score of 7.17, marking the peak performance across all tested scales.
\item \textbf{Diminishing Returns:} Further scaling to 72B leads to a slight performance drop.
\end{itemize}
These findings indicate that the 32B model offers the optimal capacity for our task, effectively leveraging our proposed frameworks.

\subsection{Human Evaluation (A/B Test)}
To assess real-world performance, we conducted a blind A/B test with 30 front-line sales professionals. These experts, chosen for their deep understanding of sales strategies and real-world business interactions, role-played as clients and engaged in hundreds of sales conversations with three AI models: our AI-Salesman, a strong SFT-only variant, and a Baseline. They then voted on paired responses, evaluating them on persuasiveness and professionalism.

The results in Table~\ref{tab:abtest_results} establish a clear performance hierarchy: Ours $\gg$ SFT-only $>$ Baseline. Our full model was preferred in 88.5\% of matchups against the baseline and 75.1\% against the strong SFT-only model. Notably, this performance ranking aligns with the results from our offline evaluations in Table~\ref{tab:main_results}, where GPT-4 served as the judge.

This quantitative strength was echoed in qualitative feedback, where evaluators praised our model for its richer, more varied language and a more natural user experience, confirming its practical value in real-world scenarios.

\begin{table}[t]
\centering
    \begin{tabular}{l ccc}
        \toprule
        \textbf{Comparison Pair} & \textbf{Win (\%)} & \textbf{Tie (\%)} & \textbf{Loss (\%)} \\
        \midrule
        Ours vs. Baseline      & \textbf{88.5} & 7.2 & 4.3 \\ % 309 25 15
        Ours vs. SFT-only      & 75.1 & 17.6 & 7.3 \\ % 268 63 26
        SFT-only vs. Baseline  & 68.7 & 21.4 & 9.9 \\ % 222 69 32
        \bottomrule
    \end{tabular}
\caption{A/B test results based on head-to-head human preference evaluations.}
\label{tab:abtest_results}
\end{table}

\section{Conclusion}
This paper introduces AI-Salesman, an end-to-end framework designed to address the limitations of Large Language Models in professional telemarketing scenarios. Our core innovations include a Bayesian-supervised reinforcement learning algorithm to optimize sales dialogue strategies directly, and the Dynamic Outline-Guided Agent mechanism for flexible, real-time conversation management.

We also constructed and released the first real-world-grounded telemarketing dataset, TeleSalesCorpus, for this task. Extensive automated and human evaluations demonstrate that our approach significantly outperforms baseline models in generating persuasive and business-compliant dialogue.

In summary, this work provides a systematic methodology and practical resources for building more effective and reliable goal-oriented persuasive AI.

\section{Acknowledgments}
We sincerely thank the reviewers for their insightful comments and valuable suggestions. This work was supported by National Key R\&D Program of China (2024YFC3308000), the Natural Science Foundation of China (No.  62476265, 62306303, 62506354), the Basic Research Program of ISCAS (Grant No. ISCAS-ZD-202401).

\bibliography{aaai2026}
% \newcommand{\isChecklistMainFile}{}
% \begingroup
% \input{ReproducibilityChecklist/LaTeX/ReproducibilityChecklist}
% \endgroup 

\onecolumn
\appendix
\setcounter{secnumdepth}{2}
\section{Related Work}
Traditional telephone sales have long faced severe challenges such as high labor costs, high employee turnover rates, and bottlenecks in conversion efficiency \citep{call_centres, ahearne2005empower, lloyd2020efficiency}. These inherent business pain points provide a clear motivation and broad application prospects for the intervention of conversational AI technology.

Early research in conversational AI primarily focused on Task-Oriented Dialogue Systems (TODS) \citep{gao2018neural, qin-etal-2023-end}. TODS typically employ a modular pipeline architecture, including components such as Natural Language Understanding, Dialogue State Tracking, Dialogue Policy, and Natural Language Generation. This architecture demonstrates high reliability in handling dialogues with clear goals and fixed processes (e.g., booking tickets, querying the weather). However, the rigidity of its design, high costs for domain extension, and vulnerability to unexpected user inputs make it difficult to meet the flexibility and persuasive skills required for telephone sales \citep{wen-etal-2017-network, feng-etal-2023-towards}.

With the advent of Large Language Models, end-to-end generative dialogue systems have become the mainstream paradigm \citep{chung-etal-2023-instructtods, dong-etal-2025-protod}. To adapt general-purpose LLMs to specific domains, the main technical paths are divided into Fine-tuning and In-Context Learning. Fine-tuning can deeply inject domain-specific knowledge into the model by updating its parameters on domain data, but this process is associated with high computational and time costs, and carries the risk of catastrophic forgetting, which may impair the model's original general capabilities \citep{kirkpatrick2017overcoming, luo2025empiricalstudycatastrophicforgetting,li-etal-2024-revisiting}. In contrast, In-Context Learning guides the model by providing task examples in the prompt, offering greater flexibility and cost-effectiveness, but the stability and depth of its knowledge injection are significantly affected by the context window length and the choice of examples, making it difficult to ensure consistency in long-process tasks \citep{NEURIPS2020_1457c0d6, liu-etal-2024-lost, agarwal2024manyshot}. RL has promoted the success of reasoning models \citep{openai2024o1, deepseekai2025deepseekr1incentivizingreasoningcapability,  chen2025xiangqir1enhancingspatialstrategic}, but its application has mainly focused on tasks with closed-form solutions like code and mathematics. How to utilize reinforcement learning to optimize multi-turn interaction strategies in Persuasive Dialogue remains an under-explored research gap.

The evaluation of generative AI dialogue systems is also a key challenge. Traditional automatic metrics based on word overlap \citep{papineni-etal-2002-bleu, lin-2004-rouge} have been shown to have a very low correlation with human judgments of open-ended dialogue quality \citep{liu-etal-2016-evaluate}.To this end, academia and industry have begun to explore new evaluation paradigms represented by LLM-as-a-Judge\citep{zheng2023judging, chan2024chateval}. This method utilizes a powerful LLM as a judge to score and evaluate the responses generated by models. Although this method shows efficiency advantages in automated evaluation, its own bias issues, the stability of evaluation results, and their consistency with real human judgments have also attracted extensive attention and in-depth research \citep{wang-etal-2024-large-language-models-fair, thakur2025judgingjudgesevaluatingalignment}. This highlights the necessity of establishing a more reliable and comprehensive evaluation system tailored to specific tasks, such as sales conversion rates.

\section{Detailed Evaluation Framework Components}
\label{app:framework_details}
This appendix provides detailed descriptions of the core components of our evaluation framework introduced in Section~\ref{sec:evaluation_framework}.

\subsection{Core Sales Capabilities}
Our framework identifies six core capabilities essential for successful telemarketing interactions. These capabilities ensure a holistic evaluation of the model performance.
\begin{itemize}
    \item \textbf{Role-playing:} This assesses the model's ability to consistently maintain a predefined persona, such as an experienced and professional account manager. The evaluation focuses on whether the model's tone, language, and conversational focus align with the specified role throughout the dialogue.
    \item \textbf{Business Analysis:} This capability measures the model's proficiency in leveraging user-specific data to deliver a personalized and persuasive sales pitch. A key aspect is the model's ability to ground its analysis strictly within the provided context, making relevant connections between the user's business status and the proposed promotional activity without hallucinating information.
    \item \textbf{Activity Introduction:} This evaluates the clarity, accuracy, and appeal of the model's presentation of the sales activity. The agent must effectively communicate all critical information, including the activity's rules, validity period, and participation methods, ensuring the user can fully comprehend the offer.
    \item \textbf{Idle-chat Rejection:} In telemarketing, maintaining focus is crucial. This capability assesses the model's skill in politely declining to engage in conversations that deviate from the sales objective. A successful model should gracefully redirect the dialogue back to the promotional activity, reinforcing its professional role and the call's purpose.
    \item \textbf{Objection Handling:} This measures the model's effectiveness in addressing and resolving user inquiries and objections. This includes clarifying ambiguities about the promotion (e.g., duration, calculating rewards) and articulating the value proposition to alleviate the user's concerns.
    \item \textbf{Operational Guidance:} This capability evaluates the model's ability to provide clear, actionable instructions that guide the user to locate and participate in the activity. This is a critical final step to convert interest into action.
\end{itemize}

\begin{table*}[h] % Use table* for a table that spans both columns
\centering
    \begingroup % Start a group to localize \arraystretch
    \begin{tabular}{p{0.25\textwidth} p{0.7\textwidth}} % Removed | for vertical lines
    \toprule
    % Adding a small vertical space before the header text
    \rule{0pt}{14pt}\textbf{Metric} & \textbf{Description}\rule{0pt}{14pt} \\ % Rule adds height above and below text
    \midrule
    Guideline Adherence & Conformance to predefined sales rules, policies, and ethical guidelines. \\
    Factual Correctness & Correctness of all presented information against the reference context. \\
    Logical Coherence & Clarity, sound reasoning, consistency, and contextual relevance of the response. \\
    User Need Fulfillment & Effectiveness in addressing the customer's explicit and implicit needs. \\
    Response Richness & Diversity, and the informativeness of the agent's responses, avoiding repetition. \\
    Safety & Absence of false promises, misleading content, or other potentially harmful content. \\
    Completeness & Coverage of all critical information points and standard operating procedures required for the dialogue turn. \\
    \bottomrule
    \end{tabular}
    \endgroup % End the group
\caption{Multi-dimensional Evaluation Metrics.}
\label{tab:multi_dimensional_metrics}
\end{table*}

\subsection{Multi-dimensional Evaluation Metrics}
To provide a fine-grained and consistent assessment across all capabilities, we evaluate each of the model's responses using a rubric of seven qualitative metrics. These metrics ensure that our evaluation is not only comprehensive but also deeply rooted in the practical requirements of a successful sales interaction. As shown in Table~\ref{tab:multi_dimensional_metrics}, the seven metrics are: Guideline Adherence, Factual Correctness, Logical Coherence, User Need Fulfillment, Response Richness, Safety, and Completeness.
\section{Detailed Data Construction Methodology}
\label{app:data_construction_details}
\newtcolorbox{AIBox}[2][]{aibox,title=#2,#1}
\tcbset{
  aibox/.style={
    width=0.95\textwidth,
    top=5pt,
    colback=black!05,
    colframe=black!20,
    colbacktitle=black!50,
    enhanced,
    center,
    attach boxed title to top left={yshift=-0.1in,xshift=0.1in},
    boxed title style={boxrule=0pt,colframe=white,},
  }
}
This appendix provides a comprehensive description of the three main stages of our data construction pipeline.

\subsection{Stage 1: Asset Distillation and Scenario Synthesis}
To ensure TeleSalesCorpus is grounded in reality, we began with a seed collection of anonymized, real-world sales conversations. Instead of using this data directly, we performed a structured analysis to distill reusable components.

\paragraph{Dialogue Flow Modeling and State-Conditioned Indexing}
To enforce a logical conversational structure, we first manually annotated our seed collection of real dialogues to model the canonical telemarketing flow: 
\texttt{Opening} $\rightarrow$ \texttt{Business\_Analysis} $\rightarrow$ \texttt{Promotion\_Introduction} $\rightarrow$ \texttt{UI\_Guidance} $\rightarrow$ \texttt{Ascertain\_Intent\_\&\_Handle\_Objections} $\rightarrow$ \texttt{Polite\_Closing}. 
For each turn in these dialogues, we created an interaction chunk consisting of a (\texttt{User\_Utterance}, \texttt{Agent\_Response}) pair and tagged each chunk with its corresponding dialogue state. Each chunk's \texttt{User\_Utterance} was then embedded and stored in a vector database, creating a state-conditioned index for targeted retrieval during simulation.

\paragraph{Scenario Synthesis}
Using insights from the real-world data, we synthesized a diverse set of dynamic business scenarios. We created a pool of 5 distinct promotional campaigns. For each dialogue to be generated, a random combination of 1-to-3 campaigns was sampled from this pool. We then used GPT-4 to author a detailed knowledge base for each promotion, guided by structured templates. This ensures every dialogue is grounded in a unique, complex, and logically consistent set of business constraints. Below is an example of a knowledge base entry.
\begin{figure}[H]
    \label{list:user-agent-prompt}
    \begin{AIBox}{}
    \parbox[t]{0.96\textwidth}{
    \small\begin{alltt}
    - [Promotion Name] Flash Recharge Bonus
    
    - [Objective] To encourage users to increase their advertising budget by offering immediate value.
    
    - [Eligibility Criteria] Users who have been online for less than 90 days.
    
    - [Pricing Tiers] Recharge 50/100, receive a 10/25 bonus coupon.
        
    - [Operational Rules] The bonus coupon is valid for 30 days and can be used for 'Keyword Bidding' and 'Homepage Banner' ads only. The coupon cannot be used to purchase other services or exchanged for cash. Limit one bonus per user during the campaign period.
    \end{alltt}}
    \end{AIBox}
    \caption{Knowledge Base Entry Example}
    \label{fig:generate_light}
\end{figure}

\subsection{Stage 2: State-Aware Three-Agent Dialogue Simulation}
We designed an LLM-mediated simulation framework involving three distinct GPT-4-powered agents: a \textbf{Sales Agent}, a \textbf{User Agent}, and a \textbf{Dialogue Manager}. (See Appendix~\ref{app:agent_prompts} for the detailed prompt of each agent).

The simulation is initiated by the User Agent. The framework then enters a turn-by-turn generation loop. Each cycle of the loop is driven by the user's reply and proceeds through the following five steps to generate the subsequent Sales Agent response:
\begin{enumerate}
    \item \textbf{User Response Generation:} The \texttt{RESPONSE} text from the Sales Agent's previous turn is sent to the User Agent. The User Agent, guided by its independent persona and the dialogue history, formulates and delivers its reply.
    
    \item \textbf{LLM-Based State Adjudication:} Immediately following the user's reply, the Dialogue Manager LLM receives the full context: the \texttt{current\_state} from the previous turn, the Sales Agent's \texttt{PROPOSED\_NEXT\_STATE}, and the User Agent's actual \texttt{response}. It analyzes this information to make a final, authoritative judgment on the true state of the conversation, which becomes the new \texttt{current\_state}.
    
    \item \textbf{State-Conditioned Retrieval:} With the dialogue state now finalized for the current turn, the Dialogue Manager takes the user's latest utterance and queries the vector database. Crucially, this search is filtered to only include interaction chunks tagged with the newly adjudicated \texttt{current\_state}.
    
    \item \textbf{Dynamic Prompt Assembly:} The Dialogue Manager assembles a new, context-rich prompt for the Sales Agent. This prompt includes the full dialogue history and is dynamically augmented with the retrieved real-world example, which serves as a style and strategy guide for that specific turn.
    
    \item \textbf{Sales Agent Response Generation:} The Dialogue Manager sends the complete prompt to the Sales Agent LLM. The Sales Agent processes this input and generates its action in the structured format: \texttt{[RESPONSE]:} \textit{\textless Your response to the user\textgreater} and \texttt{[PROPOSED\_NEXT\_STATE]:} \textit{\textless The dialogue state you intend to transition to\textgreater}.
\end{enumerate}
The \texttt{RESPONSE} generated in the final step is then delivered back to the User Agent, initiating the next cycle of the loop.

\subsection{Stage 3: Quality Assurance and Refinement}
A multi-faceted quality assurance process was implemented. First, all synthesized business scenarios and rules underwent manual review by domain experts to confirm their plausibility. After generation, we applied automated scripts to filter an initial corpus of 2500 dialogues based on several criteria:
\begin{itemize}
    \item Dialogues with fewer than 4 turns were discarded.
    \item Dialogues with high n-gram overlap between consecutive agent turns were removed.
    \item Dialogues containing unreplaced placeholder strings were filtered.
    \item Dialogues where the agent failed to mention the keywords of the sampled promotions were discarded as off-task.
\end{itemize}
Finally, for each distinct promotion type, we randomly sampled 20 full dialogues for manual review, guided by a rubric assessing coherence, realism, and strict factual faithfulness. After all filtering stages, we obtained our final, high-fidelity dataset of 2000 multi-turn conversations.

\section{Detailed Formulation of GRPO}
\label{app:appendix_grpo}

This section provides the detailed mathematical formulation for the Group Relative Policy Optimization algorithm referenced in the main text. GRPO adapts the Proximal Policy Optimization (PPO) framework \citep{schulman2017ppo}. Its key innovation is to estimate the advantage function by normalizing the rewards obtained from a group of parallel rollouts. This approach circumvents the need for an explicit value model, thereby eliminating the associated training overhead common in standard PPO implementations.

The advantage function $\mathcal{A}^{(i)}$ in GRPO is defined as the standardized measure of the $i$-th sample's reward, $R^{(i)}$, within its group. This is formalized as:
\begin{equation}
    \mathcal{A}^{(i)} = \frac{R^{(i)} - \mathbb{E}_{j \sim U(1,G)}[R^{(j)}]}{\sqrt{\mathbb{V}_{j \sim U(1,G)}[R^{(j)}] + \epsilon}}
\end{equation}
Here, $\mathbb{E}[\cdot]$ and $\mathbb{V}[\cdot]$ denote the empirical mean and variance over the set of rewards $\{R^{(j)}\}_{j=1}^G$ from the $G$ rollouts, and $\epsilon$ is a small constant for numerical stability.
This group-normalized advantage is then used to optimize the final objective function, which incorporates the clipped surrogate objective from PPO and a KL-divergence penalty term to regularize policy updates:
\begin{equation}
\begin{aligned}
&\mathcal{J}_{GRPO}(\theta) = \mathbb{E}_{q \sim P(Q), \{o_i\}_{i=1}^G \sim \pi_{\theta_{old}}(O|q)} \\ &
\frac{1}{G} \sum_{i=1}^{G} \frac{1}{|o_i|} \sum_{t=1}^{|o_i|} \Biggl\{ \min \Biggl(  \frac{\pi_{\theta}(o_{i,t}|q_i, o_{i,<t})}{\pi_{\theta_{old}}(o_{i,t}|q_i, o_{i,<t})} \mathcal{A}^{(i)}, \text{clip} \left( \frac{\pi_{\theta}(o_{i,t}|q_i, o_{i,<t})}{\pi_{\theta_{old}}(o_{i,t}|q_i, o_{i,<t})}, 1-\epsilon, 1+\epsilon \right) \mathcal{A}^{(i)} \Biggr) - \beta \mathbb{D}_{KL} \bigl[ \pi_{\theta} \| \pi_{ref} \bigr] \Biggr\}
\end{aligned}
\end{equation}

\section{Theoretical Derivation of the Bayesian-Supervised Reasoning Reward}
\label{app:bayesian_theorem}
This appendix elaborates on the theoretical foundations and step-by-step derivation of the Bayesian-Supervised Reasoning reward ($R_{\text{bayes}}$), as defined in Equation~\ref{eq:bayes_reward_main} of the main text.

\subsection{Bayes' Theorem: Theoretical Background}
Bayes' theorem is fundamental in probability theory that describes how to update the probability of a hypothesis based on new evidence. Its mathematical form is as follows:
\begin{equation}
    P(H | E) = \frac{P(E | H) P(H)}{P(E)}
\end{equation}
where:
\begin{itemize}
    \item $P(H | E)$ is the posterior probability: The probability of the hypothesis $H$ being true after observing the evidence $E$. This is the updated belief we aim to find.
    \item $P(E | H)$ is the likelihood: The probability of observing the evidence $E$ given that the hypothesis $H$ is true. It measures how well the hypothesis explains the evidence.
    \item $P(H)$ is the prior probability: The initial probability of the hypothesis $H$ being true, before considering any evidence. It represents our prior belief in $H$.
    \item $P(E)$ is the marginal likelihood of evidence: The total probability of observing the evidence $E$.
\end{itemize}
The core idea of Bayes' theorem is that the posterior is proportional to the likelihood times the prior. It provides a mathematically rigorous framework for updating our beliefs from a prior state in light of new evidence.

\subsection{Applying Bayes' Theorem to Reasoning Generation}
In our task, we map the components of Bayes' theorem as follows:
\begin{itemize}
    \item \textbf{Hypothesis $H$} is the model-generated reasoning chain $Th_t$. We hypothesize that this is a good and effective reasoning process.
    \item \textbf{Evidence $E$} is the given reference answer $A_t^*$‌. We use this evidence to evaluate the quality of our hypothesis (the reasoning chain).
\end{itemize}
Our objective is to find an optimal reasoning chain, $Th_t^{\text{optimal}}$, that maximizes the posterior probability given the reference answer $A_t^*$. This is precisely a maximum a posteriori estimation problem:
\begin{equation}
    Th_t^{\text{optimal}} = \arg\max_{Th_t} P(Th_t | A_t^*)
\end{equation}
Substituting our variables into Bayes' theorem yields:
\begin{equation}
    P(Th_t | A_t^*) = \frac{P(A_t^* | Th_t) P(Th_t)}{P(A_t^*)}
\end{equation}
When maximizing this expression with respect to $Th_t$, the denominator $P(A_t^*)$ is the probability of the reference answer, which is a constant for all candidate reasoning chains. Therefore, we can omit it from the optimization objective:
\begin{equation}
    \arg\max_{Th_t} P(Th_t | A_t^*) = \arg\max_{Th_t} P(A_t^* | Th_t) P(Th_t)
\end{equation}
The term on the right-hand side, $P(A_t^* | Th_t) P(Th_t)$, is the joint probability $P(Th_t, A_t^*)$. A high-quality reasoning chain should maximize this joint probability.

For computational convenience and numerical stability (to avoid underflow from multiplying many small probabilities), we typically optimize in log-space. Since the logarithm is a monotonically increasing function, maximizing a positive value is equivalent to maximizing its logarithm:

\begin{equation}
\begin{aligned}
\arg\max_{Th_t} \log P(Th_t, A_t^*) &= \arg\max_{Th_t}\left( \log P(Th_t) + \log P(A_t^* | Th_t) \right)
\end{aligned}
\end{equation}

Thus, we have successfully transformed the MAP problem into one of maximizing the log-joint probability. We define our reward $R_{\text{bayes}}$ as this log-joint probability, which naturally decomposes into two meaningful components:
\begin{equation}
    R_{\text{bayes}}(Th_t, A_t^*) = \underbrace{\log P(Th_t)}_{\text{Log-Prior}} + \underbrace{\log P(A_t^* | Th_t)}_{\text{Log-Likelihood}}
\end{equation}
These two components are estimated by the language model $\pi_\theta$ itself and are finally autoregressively decomposed to arrive at the computable form presented in Equation~\ref{eq:bayes_reward_main} of the main text.

\section{DOGA Framework Implementation Details}
\label{app:doga_details}

This part provides a detailed description of the two stages of the DOGA framework.

\subsection{Offline Stage: Structured Script Library Construction}
The construction of our structured script library involves a three-step process designed to distill best practices from historical data into a reusable and efficient resource.

\begin{itemize}
    \item \textbf{Data Collection and Annotation:} We begin with a corpus of historical telemarketing dialogues with high conversion rates. Using GPT-4, we programmatically classify each turn according to a predefined set of dialogue intents and annotate the user-specific information utilized (e.g., \texttt{recent\_recharge\_status}).
    \item \textbf{High-Quality Script Extraction:} We use GPT-4 to extract concise, persuasive, and generalizable scripts from the annotated dialogues that directly contribute to achieving the annotated intent.
    \item \textbf{Template Generation via Clustering and Summarization:} The extracted scripts are refined into reusable templates. We first encode scripts into vectors using \texttt{Qwen3-Embedding-0.6B} and group them using a greedy clustering algorithm (cosine similarity $>$ 0.8). Then, we employ GPT-4 to summarize each cluster into a generic template with standardized placeholders.
\end{itemize}
The final output is a structured library where templates are indexed by dialogue intent for efficient retrieval.

\subsection{Online Stage: Real-time Dialogue Management Details}

\paragraph{Constrained Dialogue Intent Classification}
Before generating a response, a lightweight classifier predicts the most appropriate dialogue intent.
\begin{itemize}
    \item \textbf{Model:} We fine-tune a Qwen2.5-7B model as our intent classifier, which takes the conversation history and previous intents as input.
    
    \item \textbf{Intent Transition Rules:} We define a finite-state machine that dictates valid transitions between intents. This constrains the classifier's prediction to a valid subset of intents based on the conversation's history, significantly improving accuracy and coherence.
\end{itemize}

\paragraph{Dynamic Prompt Assembly}
Once the intent for the current turn $I_t$ is determined, the framework assembles a tailored system prompt $P_t$. The prompt is formally composed as:
\begin{equation}
    P_t = P_{\text{static}}(H_{t-1}) \oplus D(I_t, M)
    \label{eq:dynamic_prompt}
\end{equation}
where $P_{\text{static}}$ is a base prompt containing the model's core persona and the full dialogue history up to turn $t-1$ ($H_{t-1}$), $\oplus$ denotes concatenation, and $D(I_t, M)$ is the dynamic prompt component. This dynamic part is a function of the predicted intent $I_t$ and the user profile $M$. The assembly of $D(I_t, M)$ involves:
\begin{enumerate}
    \item \textbf{Template Retrieval:} Based on the predicted intent $I_t$, corresponding templates (instructions, key points, reminders) are retrieved from the script library.
    
    \item \textbf{Personalization:} Placeholders within the retrieved template are populated with the user's specific information from their profile $M$.
\end{enumerate}
The fully assembled prompt $P_t$ is then passed to the model.

\section{Performance on Real-World Data}
\label{app:real_data_results}

To further validate the robustness of our proposed Bayesian reward, we replicated the ablation study on our held-out \textbf{Real-world Tele-sales Dataset}. This dataset, derived from anonymized expert conversations, is inherently noisier and more complex than the synthetic data.

\begin{figure}[htbp]
    \centering
    % Figure for the REAL-WORLD dataset
    \includegraphics[width=0.5\textwidth]{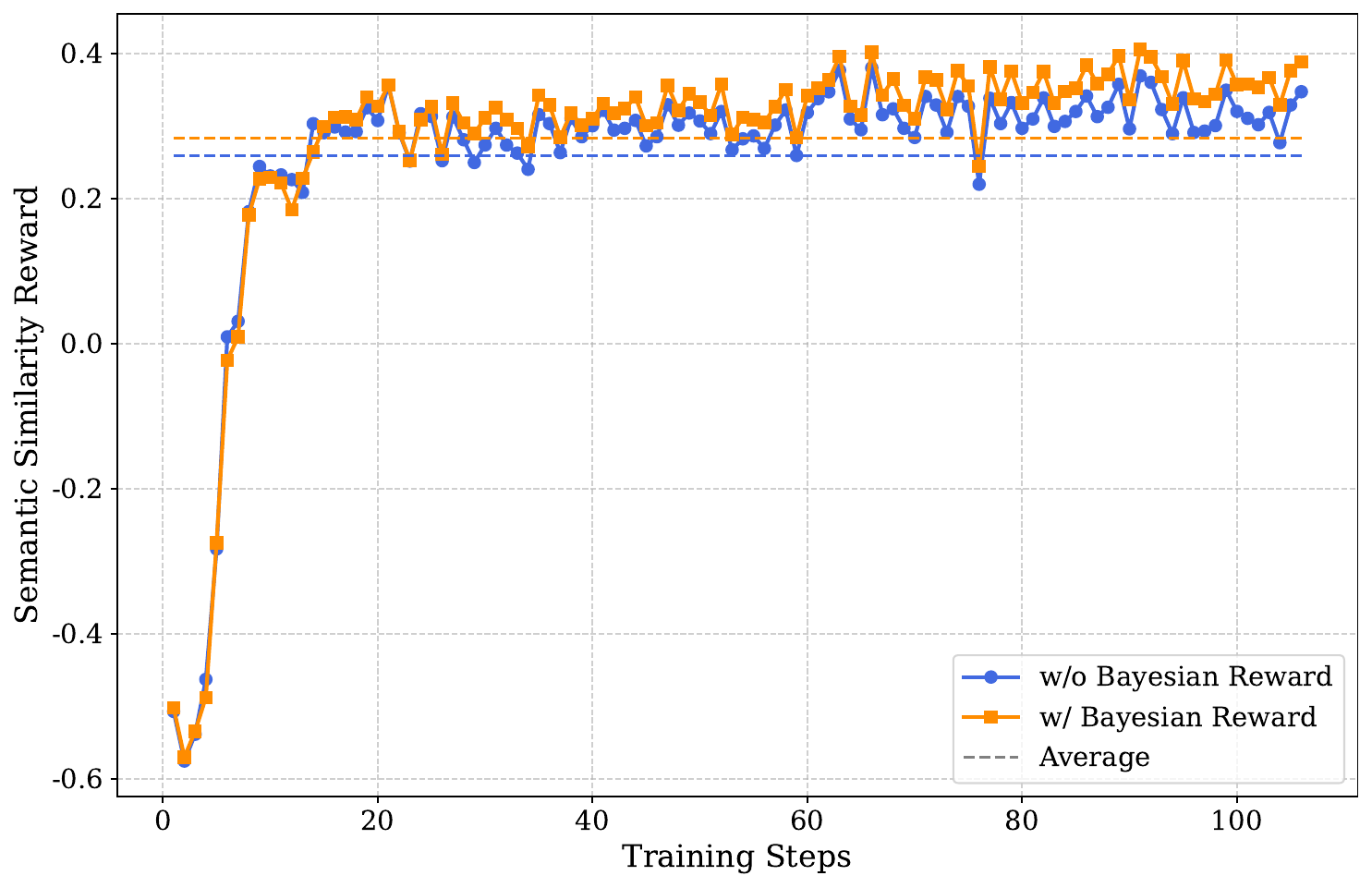}
    \caption{Performance validation on the Real-world Tele-sales Dataset. The stabilizing effect of $R_{\text{bayes}}$ remains consistent, even on more complex, non-synthetic data.}
    \label{fig:bayes_reward_effect_real}
\end{figure}

As illustrated in Figure~\ref{fig:bayes_reward_effect_real}, the performance trends are consistent with our primary findings. Although the absolute reward scores are naturally lower due to the increased difficulty of the dataset, the model trained with $R_{\text{bayes}}$ again demonstrates a markedly more stable learning curve and achieves a higher final convergence point. This confirms that the benefits of supervising the internal thought process via $R_{\text{bayes}}$ are not limited to controlled, synthetic scenarios but also translate effectively to the challenges of real-world conversational data.

\section{Agent Prompts}
\label{app:agent_prompts}
\begin{figure}[H]
    \label{list:user-agent-prompt}
    \begin{AIBox}{}
    \parbox[t]{0.96\textwidth}{
    \small\begin{alltt}
    \#\#\# Your Persona 
    
    You are the owner of "The Corner Bistro". You are busy and practical. You are open to good ideas but are very careful with your budget because your business is new. You are currently worried about the large number of competing Italian restaurants in your neighborhood.

    \#\#\# Your Task 
    
    Respond naturally to the sales agent. Raise objections based on your persona, especially concerning cost and effectiveness.
    \end{alltt}}
    \end{AIBox}
    \caption{User Agent Prompt Example}
    \label{fig:generate_light}
\end{figure}

\begin{figure}[H]
    \begin{AIBox}{}
    \parbox[t]{0.96\textwidth}{
    \small\begin{alltt}
    \#\#\# Task 
    
    Analyze the conversation snippet and determine the true dialogue state. The Sales Agent attempted to move the conversation to [PROPOSED\_NEXT\_STATE]. Based on the [USER\_RESPONSE], did the transition succeed? Choose the most accurate next state from the available list.

    \#\#\# Available States 
    
    - Opening
    
    - Business\_Analysis
    
    - Promotion\_Introduction
    
    - UI\_Guidance
    
    - Ascertain\_Intent\_\&\_Handle\_Objections
    
    - Polite\_Closing
    
    \#\#\# Context
    
    Current State: Business\_Analysis 
    
    Agent's Proposed Next State: Promotion\_Introduction
    
    User's Actual Response: ``Wait, before that, I have another question about my business analysis. You said my click-through rate was low. What can I do about that specifically?"
    
    \#\#\# Output 
    
    Business\_Analysis

    \end{alltt}}
    \end{AIBox}
    \caption{Dialogue Manager Prompt Example}
    \label{fig:generate_light}
\end{figure}

\begin{figure}[H]
    \label{list:sales-agent-prompt}
    \begin{AIBox}{}
    \parbox[t]{0.96\textwidth}{
    \small\begin{alltt}
    \#\#\# Role and Task
    
    You are a senior sales consultant. You are professional, patient, and an expert in helping new restaurant owners succeed.
    
    Your task is to generate a response to the user's last message. After crafting your response, you must also determine the most logical next state for the conversation from the available options. Your response should naturally lead the conversation into the state you propose.
    
    \#\#\# Available Dialogue States
    
    - Opening
    
    - Business\_Analysis
    
    - Promotion\_Introduction
    
    - UI\_Guidance
    
    - Ascertain\_Intent\_\&\_Handle\_Objections
    
    - Polite\_Closing
    
    \#\#\# Current Dialogue State
    
    Business\_Analysis
    
    \#\#\# Dialogue History
    
    User: "Business has been a bit slow since we opened. There are a lot of other Italian places around here, so it's hard to get noticed."
    
    \#\#\# Style Guidance for THIS TURN
    
    Emulate the style and strategy of the AGENT in the following real-world example, which was retrieved because it is highly relevant to the user's last message and the current Business\_Analysis stage:
    
    User: "We just opened, so things are still a bit slow."
    
    AGENT: "Understood. That's very common for new shops. Have you had a chance to look at your customer traffic data in the app yet? That can give us a good baseline."
    
    \#\#\# Domain Knowledge for THIS CALL
    
    You must strictly adhere to the following information. Do not mention promotions the user is not eligible for.
    
    Promotion 1: "Flash Recharge Bonus"**
    
    - Objective: To encourage users to increase their advertising budget by offering immediate value.
    
    - Eligibility Criteria: Users who have been online for less than 90 days and have an average daily ad spend of less than \$10.
    
    - Pricing Tiers: Recharge \$50/\$100, receive a 10/25 bonus coupon."
     
    - Operational Rules: "The bonus coupon is valid for 30 days and can be used for 'Keyword Bidding' and 'Homepage Banner' ads only. Limit one bonus per user."
    
    Promotion 2: "New Customer Welcome Offer"
    
    - Objective: To help new users attract their first set of customers with a compelling discount.
    
    - Eligibility Criteria: Users who have been online for less than 30 days.
    
    - Offer Details: "The platform will sponsor a 20\% off your first order coupon for your store. The cost is fully covered by the platform for the first 50 redemptions. This offer is displayed prominently to users browsing your area."
    
    - OperationalRules: "The offer runs for 14 days after activation. No cost to the user."
    
    \#\#\# User Profile
    
    - business\_name: "The Corner Bistro"
    
    - category: "Italian Restaurant"
    
    - time\_since\_onboarding: "15 days"
    
    - recent\_ad\_spend:** "\$5"
    
    - synthesized\_pain\_point: "High competition in the area; struggling to stand out."
    
    \#\#\# Behavioral Guardrails
    
    - You must not invent any features, prices, or rules not explicitly listed in the Domain Knowledge Base.
    
    - If you do not know the answer to a question, state that you will find out and get back to them.
    
    - Maintain a polite, empathetic, and helpful tone. Do not be pushy.
    
    \#\#\# Output Format
    
    You must generate your output in the following JSON format, and nothing else:
    
    \{
    
      "RESPONSE": "<Your response to the user>",
    
      "PROPOSED\_NEXT\_STATE": "<Your choice for the next dialogue state from the available list>"
    
    \}
    \end{alltt}}
    \end{AIBox}
    \caption{System Prompt for Sales Agent Example (Turn-Specific)}
    \label{fig:generate_light}
\end{figure}

\section{Experimental Settings}
\label{app:exp_settings}

\begin{table}[h!]
\centering
    \begin{tabular}{l c c c c}
    \toprule
    \textbf{Parameter} & \textbf{7B} & \textbf{14B} & \textbf{32B} & \textbf{72B} \\
    \midrule
    \multicolumn{5}{l}{\textit{\textbf{Training Configuration}}} \\
    Precision & BF16 & BF16 & BF16 & BF16 \\
    Epochs & 2 & 2 & 2 & 2 \\
    Num Generations & 4 & 4 & 4 & 4 \\
    Max Completion Length & 128 & 128 & 128 & 128 \\
    Reward Weights & 1,1,5,7 & 1,1,5,7 & 1,1,5,7 & 1,1,5,7 \\
    Global Batch Size & 256 & 256 & 256 & 256 \\
    Learning Rate (LR) & $5 \times 10^{-6}$ & $5 \times 10^{-6}$ & $5 \times 10^{-5}$ & $5 \times 10^{-6}$ \\
    Warmup Ratio & 0.1 & 0.1 & 0.1 & 0.1 \\
    DeepSpeed ZeRO Stage & 2 & 3 & 3 & 3 \\
    \midrule
    \multicolumn{5}{l}{\textit{\textbf{Hardware Configuration}}} \\
    Num GPUs(80 GB) & 8 & 8 & 32 & 32 \\
    \midrule
    \multicolumn{5}{l}{\textit{\textbf{Resource Utilization}}} \\
    Peak GPU Memory Usage & 95\% & 95\% & 95\% & 85\% \\
    Total Training Time (h) & 4 & 12 & 20 & 26 \\
    \bottomrule
    \end{tabular}
\caption{Detailed hyperparameters and resource utilization for scalability experiments.}
\label{tab:hyperparameters}
\end{table}

\end{document}